\documentclass{article}
\usepackage{spconf,amsmath,graphicx}
\usepackage{booktabs}   
\usepackage{multirow}   
\usepackage{cite}
\usepackage{amsmath,amssymb,amsfonts}
\usepackage{algorithmic}
\usepackage{textcomp}
\usepackage{xcolor}
\usepackage{epstopdf} 
\usepackage{hyperref}

\title{MELDAE: A Framework for Micro-Expression Spotting, Detection, and Automatic Evaluation in In-the-Wild Conversational Scenes}
\name{
\begin{tabular}[t]{c} 
Yigui Feng$^{1}$,
Qinglin Wang$^{1}$\sthanks{Corresponding author.},
Yang Liu$^{2}$,
Ke Liu$^{1}$,
Haotian Mo$^{1}$ \\ 
Enhao Huang$^{3}$\sthanks{Corresponding author.},
Gencheng Liu$^{1}$,
Mingzhe Liu$^{4}$,
Jie Liu$^{1}$
\end{tabular}
}
\address{ $^{1}$ College of Computer Science, National University of Defense Technology, China \\ $^{2}$ Shien-Ming Wu School of Intelligent Engineering, South China University of Technology, China \\
$^{3}$ College of Computer Science and Technology, Zhejiang University, China\\
$^{4}$ College of Computer Science and Software Engineering, Shenzhen University,  China. }
\begin{document}
\sloppy
\maketitle
\begin{abstract}
Accurately analyzing spontaneous, unconscious micro-expressions is crucial for revealing true human emotions, but this task remains challenging in ``wild'' scenarios, such as natural conversation. Existing research largely relies on datasets from controlled laboratory environments, and their performance degrades dramatically in the real world. To address this issue, we propose three contributions: the first micro-expression dataset focused on conversational-in-the-wild scenarios; an end-to-end localization and detection framework, MELDAE; and a novel boundary-aware loss function that improves temporal accuracy by penalizing onset and offset errors. Extensive experiments demonstrate that our framework achieves state-of-the-art results on the WDMD dataset, improving the key $F1_{DR}$ localization metric by 17.72\% over the strongest baseline, while also demonstrating excellent generalization capabilities on existing benchmarks.
\end{abstract}
\section{Introduction}
\label{sec:intro}

Micro-expressions (MEs), as involuntary facial muscle movements characterized by their extremely brief duration and subtle amplitude, are widely regarded as reliable cues for revealing an individual's true emotions that they attempt to suppress or conceal\cite{Ekman1969}. However, despite the considerable advancements in micro-expression analysis over the past decades\cite{Oh2018,Zhang2021}, the vast majority of research has heavily relied on data collected in strictly controlled laboratory environments. While such controlled settings simplify the complexity of data acquisition and analysis, they also deviate significantly from the complexities of real-world scenarios. "In-the-wild" settings, particularly dynamic and interactive conversational contexts, introduce a series of formidable challenges, including unconstrained head poses, variable illumination conditions, and significant facial interference from speech-related articulations, all of which lead to a drastic degradation in the performance of existing models\cite{Chen2025}.

Among all real-world scenarios, the analysis of micro-expressions within conversational contexts is particularly crucial and unique. According to the Truth-Default Theory, humans tend to operate on a default state of belief during communication, especially in everyday interactions lacking high-stakes motives\cite{Levine2014}. This cognitive bias renders deception or concealed emotions more difficult to discern. The authenticity of micro-expressions is directly correlated with an individual's stakes and cognitive load; as a form of unconscious "emotional leakage," they can become a critical key to discerning a person's true intentions precisely within these low-risk, high-trust interactions. Therefore, the analysis of micro-expressions in natural conversations holds irreplaceable theoretical and practical value.

Nevertheless, a significant gap exists in the current field: nearly all publicly available micro-expression datasets are collected in laboratory settings using induction paradigms, rendering them inherently non-conversational and non-interactive\cite{Merghani2018}. This data-level deficiency severely impedes the transition of micro-expression research from laboratory settings to real-world applications\cite{Zhao2023}. To bridge this gap, we first construct the first micro-expression dataset focused on in-the-wild conversational scenarios (WDMD). Furthermore, recognizing that manual annotation of micro-expressions in long videos is exceedingly time-consuming, we propose an end-to-end model for Micro-Expression Localization and Detection (MELDAE), aiming to provide efficient and automated analysis capabilities for in-the-wild conversational video data. Specifically, given the inherently transient nature and ambiguous temporal boundaries of micro-expressions, precise temporal localization constitutes the core challenge of this task. To this end, we have specifically designed a novel loss function to address this challenge. 

This paper’s key contributions are three fold: WDMD Dataset: We constructed the first in-the-wild conversational micro-expression dataset, providing a crucial data foundation for research in authentic interactive environments; MELDAE Framework: We proposed an effective end-to-end framework for localizing and detecting micro-expressions in noisy, real world videos; Boundary-Aware Loss(BAL): We designed a novel loss function that significantly improves localization precision by reinforcing the learning of temporal start and end boundaries.

\begin{figure*}[t]          
  \centering
  \includegraphics[width=0.8\textwidth]{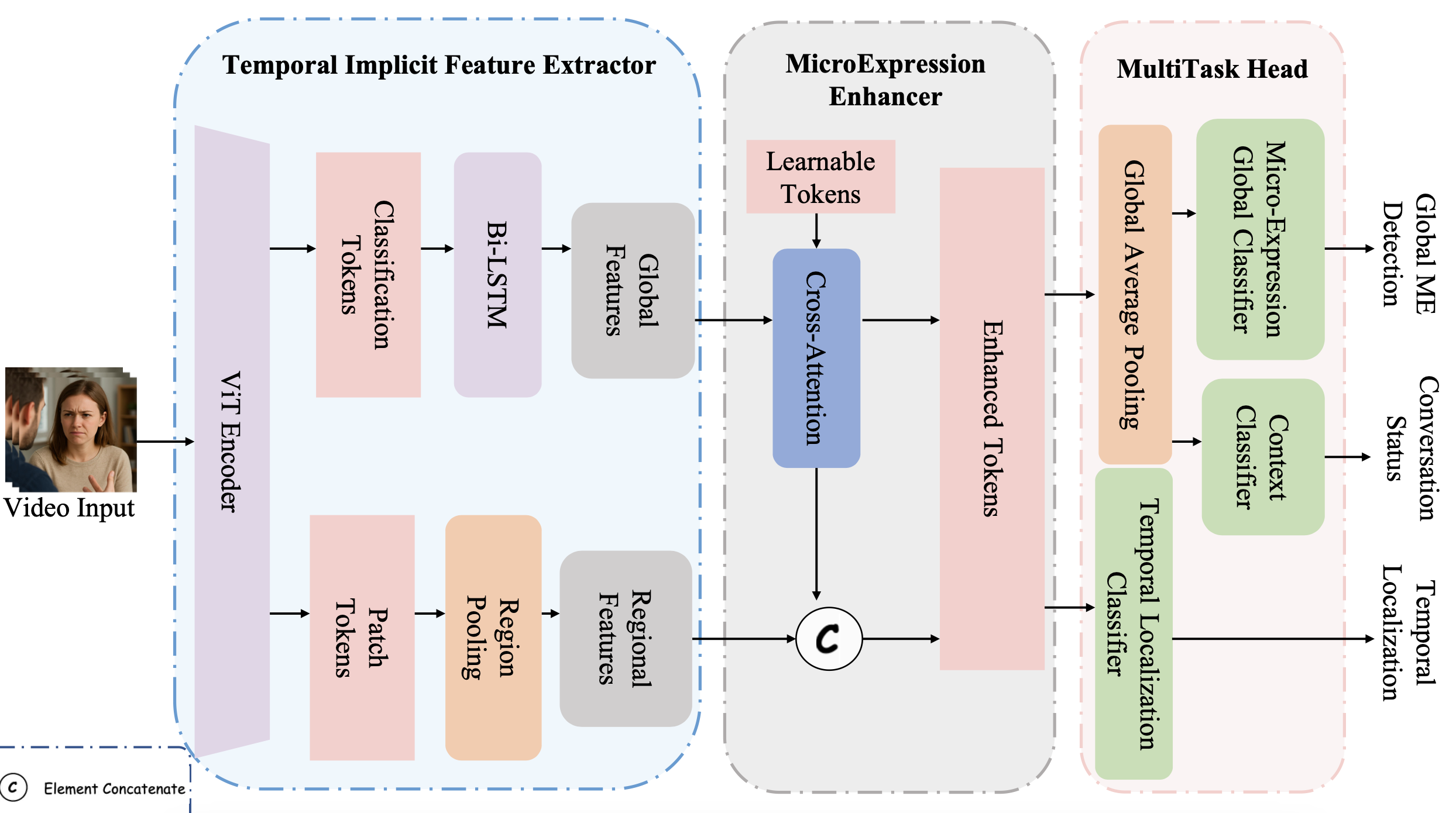}
  \caption{Overall framework of the WDMD model.}
  \label{fig1}
\end{figure*}

\section{Method}
\subsection{Model Overview}
To address the challenges of micro-expression analysis in in-the-wild conversational scenarios, we propose MELDAE, a novel end-to-end deep learning framework. As illustrated in Fig.\ref{fig1}, MELDAE follows an "encoder-enhancer-decoder" paradigm to progressively refine and analyze micro-expression signals from noisy video streams. This section details the framework's three core modules: the Temporal Implicit Feature Extractor, the MicroExpression Enhancer, and the Multitask Head. We place special emphasis on our proposed BAL function and corresponding evaluation metrics, which are crucial for achieving precise temporal localization.

\textbf{Temporal Implicit Feature Extractor:}
To handle the variable visual conditions of in-the-wild scenarios, our feature extractor employs a hierarchical strategy to robustly capture both spatial details and long-range temporal dependencies. First, for spatial representation encoding, we use a pre-trained Vision Transformer (ViT) as the frame-level encoder. Unlike CNNs, ViT's self-attention mechanism is more effective at capturing the long-range dependencies between image patches, which is crucial for modeling the coordinated muscle movements in micro-expressions. For an input video clip $X \in \mathbb{R}^{T \times C \times H \times W}$, ViT processes each of the $T$ frames to extract deep features, yielding both a global classification token and a set of local patch tokens containing rich texture information.

Following the spatial encoding, we model the temporal dimension. The sequence of classification tokens from the ViT is fed into a Bidirectional Long Short-Term Memory (Bi-LSTM) network. The Bi-LSTM captures contextual information from both past and future frames to generate a comprehensive temporal feature, $F_{\text{global}} \in \mathbb{R}^{T \times D}$, that encapsulates global motion patterns. Concurrently, the patch tokens are processed via Region Pooling to form regional features, $F_{\text{regional}}$, which retain local spatial information. Together, $F_{\text{global}}$ and $F_{\text{regional}}$ constitute the initial spatiotemporal encoding of the video content.

\textbf{MicroExpression Enhancer:}
Micro-expression signals are characterized by their high degree of sparsity and subtlety in the spatiotemporal domain, making them susceptible to being overwhelmed by large-magnitude motions such as speech articulations. To address this challenge, we have designed an attention-based enhancement module to accentuate these critical signals. The core of this module is the introduction of a set of learnable micro-expression query tokens. These tokens can be conceptualized as abstract "prototypes" or "probes" for micro-expressions, which learn to represent typical micro-expression patterns during the training process.

We leverage a cross-attention mechanism to facilitate information interaction between these query tokens and the regional features, $F_{\text{regional}}$, extracted in the previous stage. Specifically, the learnable tokens serve as the Query, while the regional features of the video simultaneously serve as both the Key and the Value. Through this mechanism, the query tokens can adaptively aggregate the most relevant features from all spatial regions across all video frames that correspond to their own learned patterns. This process effectively filters out irrelevant noise and forms a set of significantly enhanced micro-expression representations, denoted as $F_{\text{enhanced}}$. Finally, we concatenate the enhanced features, $F_{\text{enhanced}}$, with the global temporal features, $F_{\text{global}}$, to provide a more informative and discriminative feature input for the downstream tasks.

\textbf{MultiTask Head:}
The Prediction Head receives the fused, enhanced features and decodes them into outputs for three specific tasks through three parallel, structurally simple branches. The Global Micro-Expression Classifier performs Global Average Pooling on the features along the temporal dimension, which are then passed through a fully connected layer with a Sigmoid activation function to output the probability, $p_{\text{ME}} \in [0,1]$, that the entire clip contains a micro-expression. Similarly, the Conversational Scene Classifier outputs the probability, $p_{\text{State}} \in [0,1]$, that the subject is in a "speaking" state. Finally, the Temporal Locator directly processes the enhanced temporal features, $F_{\text{enhanced}}$, on a frame-by-frame basis through a fully connected layer to generate a time series $S_{\text{loc}} \in \mathbb{R}^T$, where each element $S_{\text{loc}}(t)$ represents the confidence score for the presence of a micro-expression at frame $t$.

\textbf{Boundary-Aware Multi-Task Learning Strategy:}
To jointly optimize the three tasks, we employ a weighted composite loss function:
\begin{equation}
    \mathcal{L}_{\text{total}} = w_1 \mathcal{L}_{\text{ME}} + w_2 \mathcal{L}_{\text{State}} + w_3 \mathcal{L}_{\text{loc}}
\end{equation}
where $w_1, w_2,$ and $w_3$ are task-balancing hyperparameters. For the classification losses ($\mathcal{L}_{\text{ME}}$ and $\mathcal{L}_{\text{State}}$), we use Focal Loss to counteract the severe class imbalance inherent in micro-expression data.

The core of our contribution lies in the temporal localization loss, $\mathcal{L}_{\text{loc}}$, for which we designed a BAL function. BAL addresses the critical challenge of localizing the ambiguous onset and offset boundaries of micro-expressions and is formulated as a linear combination of two components:
\begin{equation}
    \mathcal{L}_{\text{loc}} = \mathcal{L}_{\text{overlap}} + \lambda \mathcal{L}_{\text{boundary}}
\end{equation}
The Overlap Loss ($\mathcal{L}_{\text{overlap}}$) is implemented using Focal Tversky Loss. This variant of the Tversky index maximizes the overlap between prediction and ground truth while focusing on hard-to-segment samples, making it ideal for small targets with indistinct boundaries.

The key component, the Boundary Loss ($\mathcal{L}_{\text{boundary}}$), directly targets localization precision. It utilizes a weighted binary cross-entropy (BCE), $w_i \cdot \text{BCE}(p_i, y_i)$, where the weight $w_i$ is significantly increased to a value of $W_{\text{boundary}}$ for annotated start and end frames, and is otherwise set to 1. The hyperparameter $\lambda$ controls the influence of this boundary-specific penalty. This design compels the model to learn not only the ``body'' of a micro-expression but also its transient boundaries, thereby fundamentally enhancing localization performance.

\subsection{Evaluation Framework and Metrics}
To quantitatively evaluate our framework, we established a multi-dimensional metric system. For the binary classification tasks of global micro-expression detection and conversational state classification, we use standard Accuracy. For the more challenging core task of temporal localization, we employ an evaluation scheme based on Intersection over Union (IoU), where a predicted segment is deemed a True Positive (TP) if its IoU with a ground-truth segment exceeds a threshold of $\theta=0.5$. Based on this, we compute Precision and Recall to derive the primary localization metric, the F1-score. To further account for the differing facial dynamics in conversation, we separately calculate localization F1-scores for speaking ($F1_{\text{speaking}}$) and listening ($F1_{\text{listening}}$) contexts and propose a single, comprehensive metric, the F1-score for Dialogue Roles (F1$_{DR}$), defined as their harmonic mean:
\begin{equation}
    F1_{DR} = \frac{2 \cdot F1_{\text{listening}} \cdot F1_{\text{speaking}}}{F1_{\text{listening}} + F1_{\text{speaking}}}
\end{equation}
This fused metric provides a fair and robust assessment of localization performance across distinct conversational states.

\section{EXPERIMENTS AND RESULTS}
\textbf{Datasets:}
\textbf{WDMD}: a novel dataset we constructed from unscripted film conversations to capture realistic "in-the-wild" scenarios. It contains 2,253 clips (545 micro-expressions) at 60 fps and 2560×1440 resolution, featuring challenges like varied head poses, complex lighting, and speech interference. Critically, expert psychologists annotated precise onset/offset frames and the conversational context (speaking/listening). \textbf{CAS(ME)$^2$\cite{Qu2017}}:It contains 357 samples (57 micro-expressions) elicited in a controlled laboratory environment. 

\begin{figure}[t]
\begin{minipage}[b]{1.0\linewidth}
  \centering
  \centerline{\includegraphics[width=0.95\textwidth]{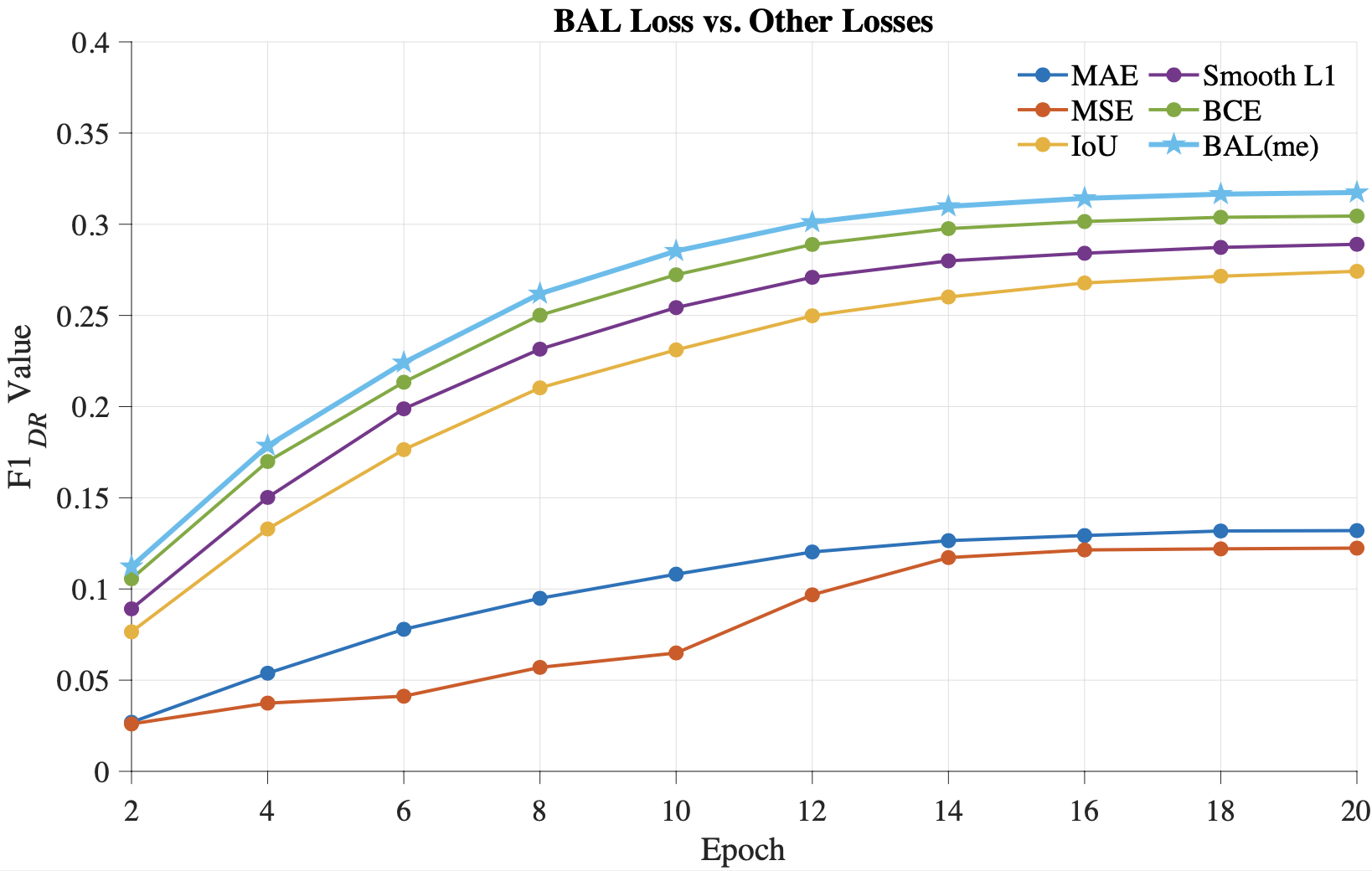}}
\end{minipage}
\caption{Experimental comparison results of BAL and other losses.}
\label{fig2}
\end{figure}

\begin{table*}[t!]
\caption{Experimental comparison of MELDAE with the baseline model. MELDAE* is the ablation model with the MicroEx-
pression Enhancer module removed from MELDAE.}
\vspace{-4mm}
\begin{center}
 
\resizebox{0.75\linewidth}{!}{
\begin{tabular}{@{}l|ccc|ccc@{}}
\toprule
\multirow{2}{*}{\textbf{Model}}              & \multicolumn{3}{c|}{\textbf{WDMD}}                                                                                                & \multicolumn{3}{c}{\textbf{CAS(ME)$^2$ }}                                                                                                   \\
                                              & \textbf{Acc\_ME} & \textbf{Acc\_State}  & \textbf{F1\textsubscript{DR}}  & \textbf{Acc\_ME}  & \textbf{F1\_Score}   \\ \midrule
LBP-TOP\cite{LeNgo2016}                                   & 0.2391                                            & 0.1980                                            & 0.0362                           & 0.3418                                          & 0.0592                                            \\
LBP-SIP\cite{Wang2014}                                   & 0.2517                                                         & 0.2203                                      & 0.0390                          & 0.3670      & 0.0733                                                                    \\ 
LTR3O\cite{Zhu2025}      & 0.7627                      & 0.7584                                       & 0.1282                                      & 0.7657                        & 0.1373                                                                             \\
CMNET \cite{Wei2023} & 0.7480                        & 0.7297                                       & 0.1016                                     & 0.7426                         & 0.1089                                                                            \\
u-BERT \cite{Nguyen2023}    & -                         & -                                      & 0.1402                                      & -                        & 0.1541                                                                         \\
PLMaM-Net \cite{Wang2024}     & 0.6907 & 0.6674                                     & 0.0825                                       & 0.7020                          & 0.0814                                                                             \\
SRMCL \cite{Bao2024}    & 0.7002                        & 0.6911                                       & 0.0807                                       & 0.7274                        & 0.0898                                                                             \\
FFDIN  \cite{Li2024}       & 0.7578                         & 0.7459                                       & 0.1130                                       & 0.7599                         & 0.1277                                                                          \\ \midrule
\textbf{MELDAE*}                        & 0.7805                 & 0.7659                             & 0.1887                             & 0.8079                & 0.2068  \\   
\textbf{MELDAE}                        & \textbf{0.8176}                & \textbf{0.8027}                              & \textbf{0.3174}                              & \textbf{0.8418}                & \textbf{0.3793} 
\\ \bottomrule
\end{tabular}}
\label{table1}
\end{center}
\vspace{-7mm}
\end{table*}

\textbf{Implementation Details:}
Our model is implemented in PyTorch and trained on eight NVIDIA H100 80GB GPUs with the AdamW optimizer. A differential learning rate strategy is applied, setting the learning rate for the ViT backbone to $1 \times 10^{-5}$ and for the newly initialized parts to $5 \times 10^{-6}$.

\textbf{Baseline Models:}
As shown in Table \ref{table1}, we present the following metrics:for the WDMD dataset, the micro-expression detection accuracy (Acc\_ME), dialogue state detection accuracy (Acc\_State), and F1-score for start and end locations (F1$_{DR}$);for the CAS(ME)$^2$ dataset, the micro-expression detection accuracy (Acc\_ME) and the average F1-Score. We also select a number of state-of-the-art methods for comparison, including manual methods (LBP-TOP \cite{LeNgo2016},LBP-SIP\cite{Wang2014}) and recent deep methods based on CNN and Transformer (LTR3O\cite{Zhu2025}, CMNET\cite{Wei2023}, u-BERT\cite{Nguyen2023}, PLMaM-Net\cite{Wang2024}, SRMCL\cite{Bao2024}, FFDIN\cite{Li2024}).

\textbf{Main Results:}
As shown in Table \ref{table1}, MELDAE framework demonstrates superior performance on the WDMD dataset, outperforming all baseline models across all metrics. This advantage is most evident in our key conversation-oriented metric, F1$_{DR}$, where MELDAE achieves a significant improvement of over 17.72\% over the strongest baseline, u-BERT. Furthermore, MELDAE's multi-task learning architecture demonstrates a unique advantage in conversation scene classification, achieving 80.27\% accuracy. To assess the model's generalization capabilities and ensure it does not overfit to the specific challenges of WDMD, we evaluated its performance on the controlled CAS(ME)$^2$ dataset. Results show that MELDAE achieves highly competitive localization F1 scores, even surpassing several methods specifically tailored for this laboratory-based benchmark. This demonstrates that MELDAE learns intrinsic and robust spatiotemporal characteristics of micro-expressions, rather than simply adapting to scene-specific noise, thus confirming its excellent generalization capabilities.

\textbf{Ablation Results:}
To isolate and validate the effectiveness of our proposed Boundary-Aware Loss, we conducted an ablation study. On the WDMD dataset, we kept the MELDAE model architecture fixed, varying only the loss function used for the temporal localization task. We compared the performance of BAL against several commonly used loss functions for classification, segmentation, and localization, including Binary Cross Entropy\cite{Wu2021}, Smooth L1\cite{Girshick2015}, IoU Loss\cite{Rezatofighi2019}, Mean Squared Error\cite{Hastie2009}, and Mean Absolute Error\cite{Bishop2006}. As illustrated in Fig.\ref{fig2}, the training curve for BAL is consistently positioned above the others and is the first to reach convergence. Moreover, the final converged F1$_{DR}$ value it achieves is also the highest among all tested loss functions.

\section{Conclusion and future work}
This paper contributes to micro-expression analysis in natural conversations across three areas: data, models, and loss functions. We built the first dedicated dataset, proposed an end-to-end detection framework, and designed a Boundary-Aware Loss function to improve localization accuracy. Extensive experiments have validated the effectiveness of our approach, and future work will focus on expanding dataset diversity and exploring lightweight models for real-time applications on mobile devices.

\bibliographystyle{IEEEbib}
\bibliography{main}

\end{document}